\documentclass[11pt, a4paper, copyright, nonumbering]{mll_style}
\usepackage{color}
\usepackage{epsfig}
\usepackage{graphicx}

\usepackage{booktabs}  %
\usepackage{tabularx}               %
\newcolumntype{C}{>{\centering\arraybackslash}X}
\usepackage{multirow}               %
\usepackage{diagbox}                %
\usepackage{hhline}                 %
\usepackage{color}                  %

\usepackage{booktabs}
\usepackage{extarrows}
\usepackage{makecell}
\usepackage{wrapfig}
\usepackage{colortbl}
\usepackage{tabularray}
\usepackage[most]{tcolorbox}
\tcbuselibrary{skins} %
\usepackage{tcolorbox}

\usepackage{amssymb}

\usepackage{adjustbox}
\usepackage{array}
\usepackage{floatflt}

\usepackage{bm}
\usepackage{nicefrac}
\usepackage{microtype}

\usepackage{changepage}
\usepackage{extramarks}
\usepackage{fancyhdr}
\usepackage{lastpage}
\usepackage{setspace}
\usepackage{soul}
\usepackage{xspace}

\usepackage{url}

\usepackage{enumerate}
\usepackage{enumitem}  %

\usepackage{makecell}

\usepackage{pifont} %

\usepackage{algorithm}
\usepackage{algpseudocode}
\usepackage{xcolor}

\usepackage[symbol]{footmisc}

\usepackage{scalefnt}

\usepackage{fontawesome5}

\usepackage{siunitx}

\usepackage{listings}

\newcolumntype{L}[1]{>{\raggedright\let\newline\\\arraybackslash\hspace{0pt}}m{#1}}
\newcolumntype{R}[1]{>{\raggedleft\let\newline\\\arraybackslash\hspace{0pt}}m{#1}}

\newcommand{\ignore}[1]{}

\makeatletter
\DeclareRobustCommand\onedot{\futurelet\@let@token\@onedot}
\def\@onedot{\ifx\@let@token.\else.\null\fi\xspace}

\def\eg{e.g\onedot} 
\def\ie{i.e\onedot}

\makeatother

\definecolor{MyBlue}{rgb}{0.46, 0.50, 0.61}
\definecolor{MyDarkBlue}{rgb}{0,0.08,0.8}
\definecolor{MyDarkGreen}{RGB}{45,155,45}
\definecolor{MyDarkRed}{rgb}{0.8,0.02,0.02}
\definecolor{MyOrange}{rgb}{1.0, 0.4, 0.2}
\definecolor{MyPurple}{RGB}{111,0,255}
\definecolor{MyRed}{rgb}{0.8,0.0,0.0}
\definecolor{MyGold}{rgb}{0.75,0.6,0.12}
\definecolor{MyDarkgray}{rgb}{0.66, 0.66, 0.66}
\definecolor{MyBrown}{rgb}{0.65, 0.16, 0.16}
\definecolor{MyMutedRose}{rgb}{0.58, 0.29, 0.35}
\definecolor{JiayuanColor}{rgb}{0.60,0.43,0.48}
\definecolor{erranColor}{rgb}{24, 40, 113}

\definecolor{citecolor}{HTML}{696FAD}

\newif\ifpropositionfirstitem
\propositionfirstitemtrue

\definecolor{bggray}{HTML}{F5F5F5}
\definecolor{pvdblue}{HTML}{DAE8FC}
\definecolor{RoseQuartzBg}{HTML}{F7CAC9}
\definecolor{RoseQuartz}{HTML}{F5A798}
\definecolor{Serenity}{HTML}{92A8D1}
\definecolor{OrangeRed}{rgb}{1.0, 0.27, 0.0}
\definecolor{RoyalBlue}{cmyk}{1, 0.50, 0, 0}
\definecolor{Turquoise}{HTML}{0F4C81}
\definecolor{mint}{rgb}{0.24, 0.71, 0.54}
\definecolor{green}{rgb}{0.0, 0.120, 0.0}

\newdimen\abovecrulesep
\newdimen\belowcrulesep
\makeatletter
\patchcmd{\@@@cmidrule}{\aboverulesep}{\abovecrulesep}{}{}
\patchcmd{\@xcmidrule}{\belowrulesep}{\belowcrulesep}{}{}
\makeatother

\definecolor{mybluetitle}{HTML}{4B527E} %

\definecolor{mygreen}{RGB}{0,150,0}
\definecolor{boxbackground}{HTML}{F0F7FF}  %
\definecolor{boxborder}{HTML}{D0D9E5}      %
\definecolor{accentblue}{HTML}{4A86E8}     %
\definecolor{lightblue}{HTML}{EEF3FF}  %
\definecolor{bordergray}{HTML}{CCCCCC}  %
\definecolor{headerblue}{HTML}{2C5AA0}  %

\definecolor{lavenderframe}{HTML}{E6E6FA}  %
\definecolor{lighterlav}{HTML}{F5F5FF}  %
\definecolor{codegray}{rgb}{0.5,0.5,0.5}  %
\definecolor{codepurple}{HTML}{483D8B}  %
\definecolor{backcolour}{HTML}{F5F5FF}  %
\lstdefinestyle{mystyle}{
    backgroundcolor=\color{backcolour},
    commentstyle=\color{headerblue},
    keywordstyle=\color{codepurple},
    numberstyle=\tiny\color{codegray},
    stringstyle=\color{codepurple},
    basicstyle=\ttfamily\scriptsize,
    breakatwhitespace=false,
    breaklines=true,
    captionpos=b,
    keepspaces=true,
    frame=none,
    numbersep=5pt,
    showspaces=false,
    showstringspaces=false,
    showtabs=false,
    tabsize=2
}

\definecolor{jsonkey}{RGB}{44, 130, 201}     %
\definecolor{jsonstring}{RGB}{255, 140, 0}   %
\definecolor{jsonnumber}{RGB}{34, 139, 34}   %

\lstdefinelanguage{json}{
    basicstyle=\ttfamily\small,
    numbers=left,
    numberstyle=\tiny\color{gray},
    stepnumber=1,
    numbersep=5pt,
    showstringspaces=false,
    breaklines=true,
    frame=none,
    backgroundcolor=\color{gray!5},
    literate=
     *{:}{{{\color{jsonkey}:}}}{1}
      {,}{{{\color{jsonkey},}}}{1}
      {"}{{{\color{jsonstring}"}}}{1}
      {[}{{{\color{jsonkey}[}}}{1}
      {]}{{{\color{jsonkey}]}}}{1}
      {0}{{{\color{jsonnumber}0}}}{1}
      {1}{{{\color{jsonnumber}1}}}{1}
      {2}{{{\color{jsonnumber}2}}}{1}
      {3}{{{\color{jsonnumber}3}}}{1}
      {4}{{{\color{jsonnumber}4}}}{1}
      {5}{{{\color{jsonnumber}5}}}{1}
      {6}{{{\color{jsonnumber}6}}}{1}
      {7}{{{\color{jsonnumber}7}}}{1}
      {8}{{{\color{jsonnumber}8}}}{1}
      {9}{{{\color{jsonnumber}9}}}{1}
}

\newtcblisting{jsonbox}{
  listing engine=listings,
  colback=gray!3!white,
  colframe=gray!75!black,
  boxrule=0.4mm,
  arc=2mm,
  outer arc=2mm,
  breakable,
  enhanced,
  listing only,
  listing options={language=json}
}

\newtcolorbox{promptbox}[2][]{ %
    enhanced,
    breakable,
    boxsep=5pt,
    left=9pt,
    right=7pt,
    top=5pt,
    bottom=5pt,
    colback=boxbackground,
    colframe=boxborder,
    boxrule=0.5pt,
    arc=4pt,
    frame hidden, %
    borderline west={3pt}{0pt}{accentblue},
    shadow={0.5pt}{0.5pt}{1.5pt}{black!10},
    fontupper=\normalsize,
    title=#2, %
    colbacktitle=accentblue, %
    coltitle=white,         %
    fonttitle={\fontsize{9}{11}\selectfont\bfseries}, %
    attach boxed title to top left={yshift=-2.5mm, xshift=3.2mm},
    boxed title style={
        enhanced,
        left=3pt,
        right=3pt,
        top=1pt,    %
        bottom=1pt, %
        boxsep=2pt,
        arc=3pt,
        boxrule=0pt,
        colback=accentblue,
    },
    #1 %
}

\newtcolorbox{notitlepromptbox}[1][]{
    enhanced,
    breakable,
    boxsep=5pt,          %
    left=9pt,            %
    right=7pt,           %
    top=5pt,             %
    bottom=5pt,          %
    colback=boxbackground,
    colframe=boxborder,
    boxrule=0.5pt,
    arc=4pt,             %
    frame hidden,
    borderline west={3pt}{0pt}{accentblue},  %
    shadow={0.5pt}{0.5pt}{1.5pt}{black!10},  %
    notitle,
    fontupper=\normalsize,    %
    #1
}

\newtcolorbox{onebox}[2][]{
    enhanced, 
    center title,
    left*=0pt, right*=0pt,
    boxsep=2pt, left=5pt, right=5pt,
    skin first=enhanced,
    skin middle=enhanced,
    skin last=enhanced,
    colframe = mybluetitle!90,
  colback  = mybluetitle!10,
    fonttitle=\bfseries\rmfamily\fontfamily{phv}\selectfont,
    title={\footnotesize\strut{#2}  \refstepcounter{subsubsection} \addcontentsline{toc}{subsubsection}{\string\numberline{\thesubsubsection}#2}
    },
    #1
    }

\usepackage{algorithm}          
\usepackage{algpseudocode}      
\usepackage{amsmath}            
\usepackage{amsfonts}           
\usepackage{bm}   

\usepackage{amsfonts}
\usepackage[numbers]{natbib}
\usepackage{graphicx}
\usepackage{xspace}
\usepackage{xcolor}
\definecolor{highlightgray}{RGB}{220, 220, 220}
\usepackage{adjustbox}
\usepackage{array}
\usepackage{float}
\usepackage{geometry}
\usepackage{dblfloatfix}
\usepackage{enumitem}
\usepackage{setspace}
\usepackage{booktabs}
\usepackage{multirow}
\usepackage{tabularx}
\usepackage{siunitx}
\usepackage{amsmath,array}

\usepackage{pdflscape}

\usepackage{amsmath}
\usepackage{amssymb}
\usepackage{amsfonts}
\usepackage{mathtools}
\usepackage{bm}
\usepackage{dsfont}

\usepackage{tikz}
\usepackage{pgfplots}
\pgfplotsset{compat=1.18}
\usepackage{algorithm}
\usepackage{algpseudocode}
\usepackage{listings}

\usepackage{cleveref}
\hypersetup{colorlinks=true}

\usepackage{enumitem}
\usepackage{todonotes}
\usepackage{fontawesome5}
\usepackage[most]{tcolorbox}
\usepackage{CJKutf8}
\usepackage{lipsum}

\usepackage[misc]{ifsym}
\usepackage[table]{xcolor}
\usepackage{makecell}
\usepackage{colortbl}
\usepackage{bbm}
\usepackage{wrapfig}

\definecolor{kbBlue}{RGB}{92, 131, 227}     
\definecolor{kbLight}{RGB}{237, 244, 255}   
\definecolor{kbBar}{RGB}{92, 131, 227}      
\usepackage{tcolorbox}
\tcbuselibrary{skins,breakable}

\newtcolorbox{prompt}{%
  enhanced,
  breakable,
  colback=kbLight,
  colframe=kbLight,
  boxrule=0pt,
  arc=8pt,
  left=12pt, right=12pt, top=12pt, bottom=12pt,
  borderline west={4pt}{0pt}{kbBar},
}

\usepackage{tikz}
\newcommand{\hollowcircled}[1]{%
  \tikz[baseline=(char.base)]{%
    \node[shape=circle,draw,inner sep=1pt] (char) {#1};%
  }%
}

\algdef{SE}[FORALL]{ForAll}{EndForAll}[1]{\algorithmicforall\ #1\ }{}
\algtext*{EndForAll}

\definecolor{tableblue}{RGB}{201,226,239}

\makeatletter
\def\@BTrule[#1]{%
  \ifx\longtable\undefined
    \let\@BTswitch\@BTnormal
  \else\ifx\hline\LT@hline
    \nobreak
    \let\@BTswitch\@BLTrule
  \else
     \let\@BTswitch\@BTnormal
  \fi\fi
  \global\@thisrulewidth=#1\relax
  \ifnum\@thisruleclass=\tw@\vskip\@aboverulesep\else
  \ifnum\@lastruleclass=\z@\vskip\@aboverulesep\else
  \ifnum\@lastruleclass=\@ne\vskip\doublerulesep\fi\fi\fi
  \@BTswitch}
\makeatother

\addto\extrasenglish{
}

 {\begin{list}{}%
         {\setlength{\leftmargin}{#1}}%
         \item[]%
 }
 {\end{list}}

\usepackage[toc,page,header]{appendix}
\usepackage{minitoc}

\reportnumber{001} %

\title{\centering Continual Vision-Language Learning for Remote Sensing: Benchmarking and Analysis
}

\date{}

\author[*]{
Xingxing Weng$^{1,*}$,
Ruifeng Ni$^{4,*}$,
Chao Pang$^{2,*}$,
XiangYu Hao$^{1}$,
Yishan Wang$^{5}$,
Xiaokang Zhang$^{2}$,
Wei Xu$^{3}$,
Gui-Song Xia$^{2,}\textsuperscript{\Letter}$

{\small $^*$Equal Contribution; $^{\textsuperscript{\Letter}}$Corresponding Author}
\\
\small \textsuperscript{1}School~of~Computer~Science, Wuhan~University,
\textsuperscript{2}School~of~Artificial~Intelligence, Wuhan~University,
\textsuperscript{3}Faculty~of~Geographical~Science, Beijing~Normal~University,
\textsuperscript{4}College~of~Resources~and~Environmental~Sciences, Shanxi~Agricultural~University,
\textsuperscript{5}School of Electronic Information, Wuhan~University
\\
\vspace{-10pt}
}

\begin{abstract}
  Current remote sensing vision-language models (RS VLMs) demonstrate impressive performance in image interpretation but rely on static training data, limiting their ability to accommodate continuously emerging sensing modalities and downstream tasks. This exposes a fundamental challenge: enabling RS VLMs to continually adapt without catastrophic forgetting. Despite its practical importance, the continual learning capability of RS VLMs remains underexplored, and no dedicated benchmark currently exists. In this work, we present CLeaRS, a comprehensive benchmark for continual vision-language learning in remote sensing. CLeaRS comprises 10 curated subsets with over 207k image-text pairs, spanning diverse interpretation tasks, sensing modalities, and application scenarios. We further define three evaluation protocols: long-horizon, modality-incremental, and task-incremental settings, to systematically assess continual adaptation. Extensive benchmarking of diverse vision-language models reveals catastrophic forgetting across all settings. Moreover, representative continual learning methods, when adapted to RS VLMs, exhibit limited effectiveness in handling task, instruction, and modality transitions. Our findings underscore the need for developing continual learning methods tailored to RS VLMs.

\vspace{8pt}


\textbf{Dataset}: \href{https://github.com/XingxingW/CLeaRS-Preview}{https://github.com/XingxingW/CLeaRS-Preview}\\~
\textbf{Email: } \texttt{guisong.xia@whu.edu.cn}

\end{abstract}

\begin{document}
\begin{CJK*}{UTF8}{gbsn}

\doparttoc %
\faketableofcontents %

\maketitle

\section{Introduction}

Recent years have witnessed a number of remote sensing vision-language models (RS VLMs) that bridge remote sensing images and natural language, enabling a deeper understanding of remote sensing scenes~\cite{weng2025vision,xiao2025foundation}. Most research on RS VLMs has focused on creating instruction-following datasets for supervised fine-tuning, designed to encompass a wide variety of interpretation tasks and sensing modalities~\cite{kuckreja2024geochat,zhan2025skyeyegpt,luo2024skysensegpt,zhang2024earthgpt,pang2025vhm,zhang2024earthmarker,wang2024ringmogpt}. This diversity is expected to facilitate the deployment of RS VLMs across various downstream applications. However, this expectation is difficult to fulfill in practice, as the continuous advancement of remote sensing technologies and the ever-evolving nature of application scenarios inevitably impose new requirements on RS VLMs, including the need to handle novel sensing modalities and previously unseen interpretation tasks. 

To cope with dynamically changing requirements, a common practice is to aggregate instruction data collected at different time steps for joint fine-tuning, as exemplified by TEOChat~\cite{irvin2025teochat}, which combines GeoChat~\cite{kuckreja2024geochat} instructions with newly constructed temporal instruction data. While straightforward to implement, this solution raises concerns regarding data storage and training efficiency, especially as instruction data continues to accumulate over time. This naturally leads to a fundamental question: can RS VLMs be continuously adapted using only newly emerging instruction data, so as to acquire new vision-language knowledge without forgetting previously learned capabilities? In other words, \textit{to what extent are RS VLMs capable of continual vision-language learning?}

Existing research on continual learning in the remote sensing community has predominantly focused on models designed for a single interpretation task, such as semantic segmentation~\cite{xie2024missnet,weng2025class,sun2025mitigating} or scene classification~\cite{wei2025class,lu2021lil,ye2024multiscale}. These task-specific models are typically trained on small-scale datasets and thus possess limited knowledge coverage. As a result, prior studies mainly investigate continual learning within a single interpretation task, for example, by incrementally recognizing new semantic categories (class-incremental learning)~\cite{rong2023micro,xie2024missnet,sun2025mitigating} or adapting to distribution shifts under a fixed sensing modality (domain-incremental learning)~\cite{weng2024mdinet,wang2023domain,huang2024domain}.

In contrast to task-specific models, RS VLMs are pre-trained on large-scale datasets and thus endowed with broad vision–language knowledge. By following natural language instructions, a single RS VLM can flexibly support diverse interpretation tasks. \textit{This fundamental difference renders existing continual learning paradigms inadequate for RS VLMs for two main reasons.} First, continual learning in RS VLMs inherently requires the co-adaptation of language and visual representations to accommodate new instructions and visual concepts, whereas existing studies are largely vision-only. Second, such learning has the potential to extend beyond incremental classes or unimodal data distribution shifts, encompassing complex scenarios such as adaptation to new sensing modalities, learning previously unseen interpretation tasks, and evolving instruction formats. Therefore, there is an urgent need to establish dedicated benchmarks and evaluation protocols for systematically assessing continual vision-language learning in RS VLMs.

To bridge this gap, we introduce CLeaRS, a comprehensive benchmark for investigating continual learning in RS VLMs. CLeaRS emulates the evolving remote sensing data stream encountered during model deployment, which initially supports multiple interpretation tasks within a single sensing modality, then progressively expands to diverse sensing modalities driven by the emergence of new imaging technologies, and finally focuses on specific downstream applications. Accordingly, the instruction data in CLeaRS exhibit multi-task, multi-modal, and multi-application characteristics. CLeaRS is constructed by integrating diverse datasets with automatically generated instructions, subsequently verified by experts for quality control. Specifically, it comprises 10 subsets with 207,753 image-text pairs over 93,294 unique images. These subsets collectively cover four core tasks on optical images, including scene classification, image captioning, visual grounding, and visual question answering, span SAR and infrared modalities, and further include application-oriented data for disaster management. Statistics and examples are shown in Table~\ref{tab:clears-overview} and Fig.~\ref{fig:clears-example}.

Leveraging CLeaRS, we design three evaluation protocols to systematically assess continual vision-language learning under different complex scenarios. Under each protocol, we benchmark several vision-language models (generic and remote sensing-specific) to analyze their performance. Moreover, since dedicated continual learning methods for RS VLMs remain largely unexplored, we adapt representative continual learning methods from the general domain and evaluate their effectiveness in the remote sensing context. Our benchmarking results show that RS VLMs suffer from catastrophic forgetting, influenced by task, instruction, and modality transitions, while existing continual learning methods provide limited mitigation.

The main contributions of this paper are summarized as follows:

\begin{itemize}[label=\small$\bullet$]
    \item We introduce the CLeaRS benchmark, which comprises 10 instruction subsets spanning multiple interpretation tasks, sensing modalities, and application scenarios, facilitating a systematic exploration of continual vision-language learning in RS VLMs.

    \item We establish three evaluation protocols on CLeaRS and, under these settings, benchmark diverse vision-language models to analyze catastrophic forgetting and its contributing factors.

    \item We investigate the effectiveness and limitations of representative continual learning methods from the general domain when adapted to RS VLMs, offering insights to guide future dedicated method development.
\end{itemize}

\section{Related Work}

\textbf{Continual Learning in Remote Sensing} has focused on class-incremental scene classification~\cite{wei2025class,ye2024multiscale,fu2024class} and semantic segmentation~\cite{xie2024missnet,sun2025mitigating,rong2023micro}, where catastrophic forgetting arises from missing or degraded supervision of previously learned classes. To mitigate forgetting, prior studies resort to techniques such as knowledge distillation~\cite{shan2021class,sun2025mitigating}, memory replay~\cite{fu2024class}, and network expansion~\cite{ye2024multiscale}, often with variants tailored to remote sensing images~\cite{arnaudo2022contrastive} or interpretation tasks~\cite{shan2022class,rong2023micro,wei2025class}. Evaluation is commonly conducted on vision-only remote sensing datasets, where class-incremental scenarios are simulated through class-wise dataset splits. Despite this progress, such paradigms are insufficient for RS VLMs, which requires continual acquisition of visual and linguistic knowledge, extending beyond class increments to new tasks and sensing modalities. To address this gap, we introduce CLeaRS, a benchmark tailored to RS VLMs with multi-task, multi-modal, and multi-application coverage, and along with multiple evaluation protocols for systematic assessment.

\textbf{Benchmarks for Continual Vision-Language Learning:} Beyond remote sensing, several benchmarks have been proposed to study continual vision-language learning in the general domain~\cite{chen2024coin,guo2025hide,zhao2025mllm,pian2024modality,chen2025sefe,zhang2025merge}. Representative benchmarks such as CoIN~\cite{chen2024coin}, MLLM-CL~\cite{zhao2025mllm}, and MICL~\cite{pian2024modality} target task, domain, and modality incremental learning, respectively, and have facilitated method development. However, these benchmarks are built upon natural images and general-domain instructions, limiting their applicability to RS VLMs.

\section{CLeaRS: Continual Learning Benchmark for RS VLMs}
CLeaRS consists of 10 carefully curated subsets, totaling 207,753 image-text pairs over 93,294 unique remote sensing images. As summarized in Table~\ref{tab:clears-overview}, the benchmark covers four fundamental interpretation tasks on optical images, extends to SAR and infrared modalities with grounding and question answering tasks, and further includes application-oriented instruction data for natural disaster management. Together, these subsets span tasks, modalities, and application scenarios progressively, enabling systematic investigation of continual vision-language learning behaviors in RS VLMs. To ensure the accessibility of the CLeaRS benchmark, we curate its data sources from a collection of publicly available remote sensing datasets, covering both image-only and image-text data. The selected datasets include AID~\cite{xia2017aid}, VRSBench~\cite{li2024vrsbench}, SARDet-100K~\cite{li2024sardet}, SARLANG-1M~\cite{wei2025sarlang}, DroneVehicle~\cite{sun2022drone}, FireRisk~\cite{shen2023firerisk}, and RescueNet-VQA~\cite{sarkar2023rescuenet}. From these datasets, we transform the data into CLeaRS subsets using automated tools, followed by manual verification to ensure data quality. Below, we elaborate on the design of individual CLeaRS subsets, including their construction and statistical characteristics, and subsequently define the evaluation protocols for continual learning in RS VLMs.

\begin{table}[t]
    \centering
    \caption{Dataset statistics of the CLeaRS benchmark. The image modality includes SAR (Synthetic Aperture Radar), optical, and IR (infrared). For the \#Train and \#Test columns, \textit{a (b)} indicates the number of image-text pairs and images, respectively.}
    \label{tab:clears-overview}
    \resizebox{\textwidth}{!}{
    \begin{tabular}{cccccc}
    \toprule
    Dataset & Image Modality & Interpretation Task & \#Train & \#Test & Data Source \\ \midrule
    AID       & Optical     & Scene Classification      & 8,000 (8,000)  & 2,000 (2,000) & AID \\
    VRS-Cap   & Optical     & Image Captioning          & 7,833 (7,833)  & 2,340 (2,340) & VRSBench \\
    VRS-VG    & Optical     & Visual Grounding          & 14,569 (7,812) & 4,416 (2,324) & VRSBench \\
    VRS-VQA   & Optical     & Visual Question Answering & 16,258 (4,612) & 17,018 (4,680)& VRSBench \\
    SAR-VG    & SAR         & Visual Grounding          & 9,976 (8,000)  & 2,730 (2,157) & SARDet-100K \\
    SAR-VQA   & SAR         & Visual Question Answering & 20,000 (8,000) & 5,000 (2,000) & SARLANG-1M \\
    IR-VG     & IR          & Visual Grounding          & 16,632 (8,000) & 3,924 (1,964) & DroneVehicle \\
    IR-VQA    & IR          & Visual Question Answering & 8,487 (8,000)  & 2,056 (2,000) & DroneVehicle \\
    FireRisk  & Optical     & Risk Assessment           & 7,700 (7,700)  & 2,100 (2,100) & FireRisk \\
    RescueNet & Optical     & Damage Assessment         & 42,012 (2,829)  & 14,702 (943)& RescueNet-VQA \\
    \bottomrule
    \end{tabular}}
\end{table}

\subsection{Benchmark Curation}
\begin{figure}[t]
    \centering
    \includegraphics[width=\textwidth]{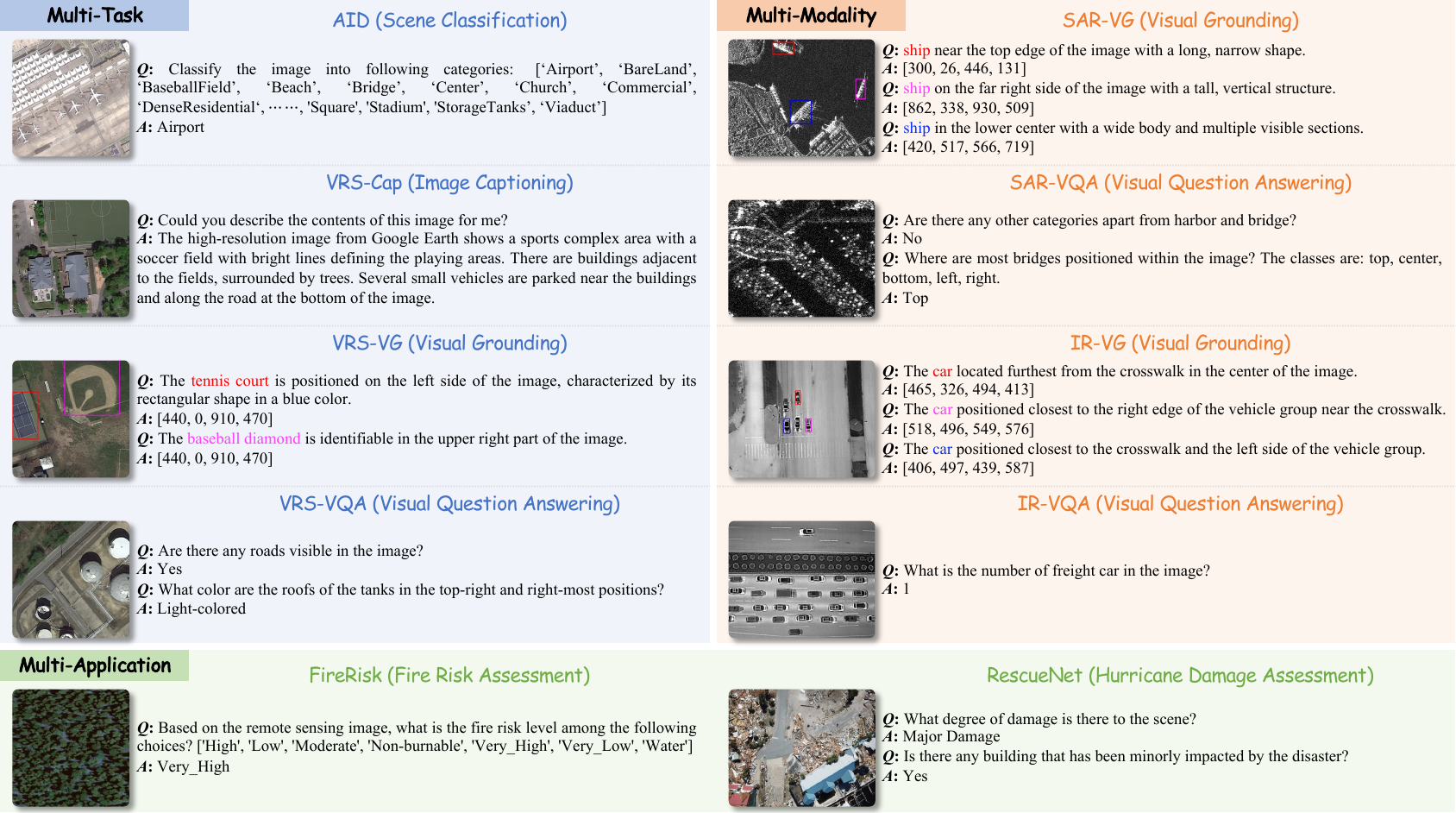}
    \caption{The CLeaRS benchmark comprises 10 subsets that progressively cover diverse interpretation tasks, sensing modalities, and application scenarios, facilitating systematic investigation of continual vision-language learning behaviors in RS VLMs.}
    \label{fig:clears-example}
\end{figure}

CLeaRS is designed to emulate remote sensing data streams encountered by RS VLMs in real-world deployment. Its subsets are organized along three dimensions: multi-task, multi-modality, and multi-application. Fig.~\ref{fig:clears-example} shows the subsets under each dimension with representative samples. 

\textbf{Multi-Task:} This dimension includes four interpretation tasks on optical images: scene classification, image captioning, visual grounding, and visual question answering. These tasks are widely supported by existing RS VLMs~\cite{pang2025vhm,zhan2025skyeyegpt,zhang2024earthgpt,muhtar2024lhrs,kuckreja2024geochat,luo2024skysensegpt}, and optical imagery is chosen for its wide availability. For scene classification, we partition the AID dataset into training and test sets with an 8:2 ratio, and transform it into instruction-following data using the template: "\textit{Classify the image into following categories: $\{$categories$\}$}", where \textit{$\{$categories$\}$} corresponds to the 30 predefined aerial scene types. The remaining three subsets are derived from VRSBench, which provides instruction-style annotations, including captions, object referring expressions, and question-answer pairs. The visual question answering subset (VRS-VQA) adopts DIOR~\cite{li2020object}-based question-answer pairs, with scene-related questions removed to avoid overlap with scene classification. The visual grounding (VRS-VG) and image captioning (VRS-Cap) subsets are constructed from DOTA-v2~\cite{ding2021object}-based image-text pairs. To prevent image overlap, images are split proportionally between the two subsets. VRS-VG samples are selected to enforce a 1:1 balance between unique and non-unique objects, and the remaining images form VRS-Cap.

\textbf{Multi-Modality:} This dimension covers non-optical imagery, including SAR and infrared data, each associated with visual grounding and visual question answering tasks. \textbf{As no publicly available grounding datasets exist for SAR or infrared images, we construct dedicated subsets} based on the SAR object detection dataset SARDet-100K and the RGB-infrared vehicle detection dataset DroneVehicle, respectively. Referring expressions are automatically generated using modality-specific prompts with Qwen3-VL~\cite{Qwen3-VL} (SAR) and Gemini-3~\cite{gemini3} (infrared). Specifically, each SARDet-100K image contains only 2.11 instances on average, leading to a sparse object distribution that facilitates unambiguous grounding. We design SAR-specific prompts instructing Qwen3-VL to leverage discriminative object attributes, including category, shape, size, and absolute/relative position, while explicitly prohibiting color-related descriptions. In contrast, infrared images in DroneVehicle are densely populated with similar objects (\eg, car) across urban roads, highways, and parking lots, making unique identification based solely on intrinsic attributes challenging. To address this, infrared-specific prompts guide Gemini-3 to describe objects using size, absolute position, and relationships with neighboring objects (\eg, next to the crosswalk or near a line of trees), ensuring unambiguous grounding. Detailed modality-specific prompts and data preprocessing procedures are provided in Appendix. The automatically generated referring expressions are further manually verified for quality control, resulting in two subsets: SAR-VG with 9,976/2,730 and IR-VG with 16,632/3,924 training/testing image-text pairs, respectively.

To support visual question answering, we construct dedicated subsets for SAR and infrared images. SAR-VQA is sampled from SARLANG-1M, which contains question-answer pairs derived from SARDet-100K images and covers five question types: object identification, classification, instance counting, positioning, and region referring. We randomly sample 20,000 training and 5,000 testing pairs. SAR-VQA images do not overlap with SAR-VG to prevent data leakage. For the infrared modality, we construct IR-VQA using DroneVehicle images and generate question-answer pairs for four question types, excluding object classification because most infrared images contain multiple annotated vehicle categories. Combining question templates with ground-truth annotations, IR-VQA contains 8,487 training and 2,056 testing pairs. Balanced question distributions and non-overlapping images with IR-VG are similarly enforced. 

\textbf{Multi-Application:} This dimension shifts from general-purpose remote sensing interpretation to application-specific scenarios, including fire risk and hurricane damage assessment. The risk assessment subset is constructed from FireRisk, where each image is labeled with one of seven fire risk levels. We cast it into an instruction format using the template: "\textit{Based on the remote sensing image, what is the fire risk level among the following choices? $\{$categories$\}$}", and uniformly sample across risk levels, yielding 7,700 training and 2,100 testing image-text pairs. For hurricane damage assessment, we adopt RescueNet-VQA, built upon aerial images collected after Hurricane Michael and covering eight question types: simple and complex counting, building and road condition recognition, level of damage, risk assessment, density estimation, and positional reasoning.

\subsection{Benchmark Statistics}
To provide a quantitative characterization of CLeaRS, we present detailed statistics for the five newly constructed subsets, namely AID, FireRisk, SAR-VG, IR-VG, and IR-VQA, while the remaining subsets are adapted from existing image-text datasets and described in Appendix. 

\begin{figure}[t]
    \centering
    \includegraphics[width=\textwidth]{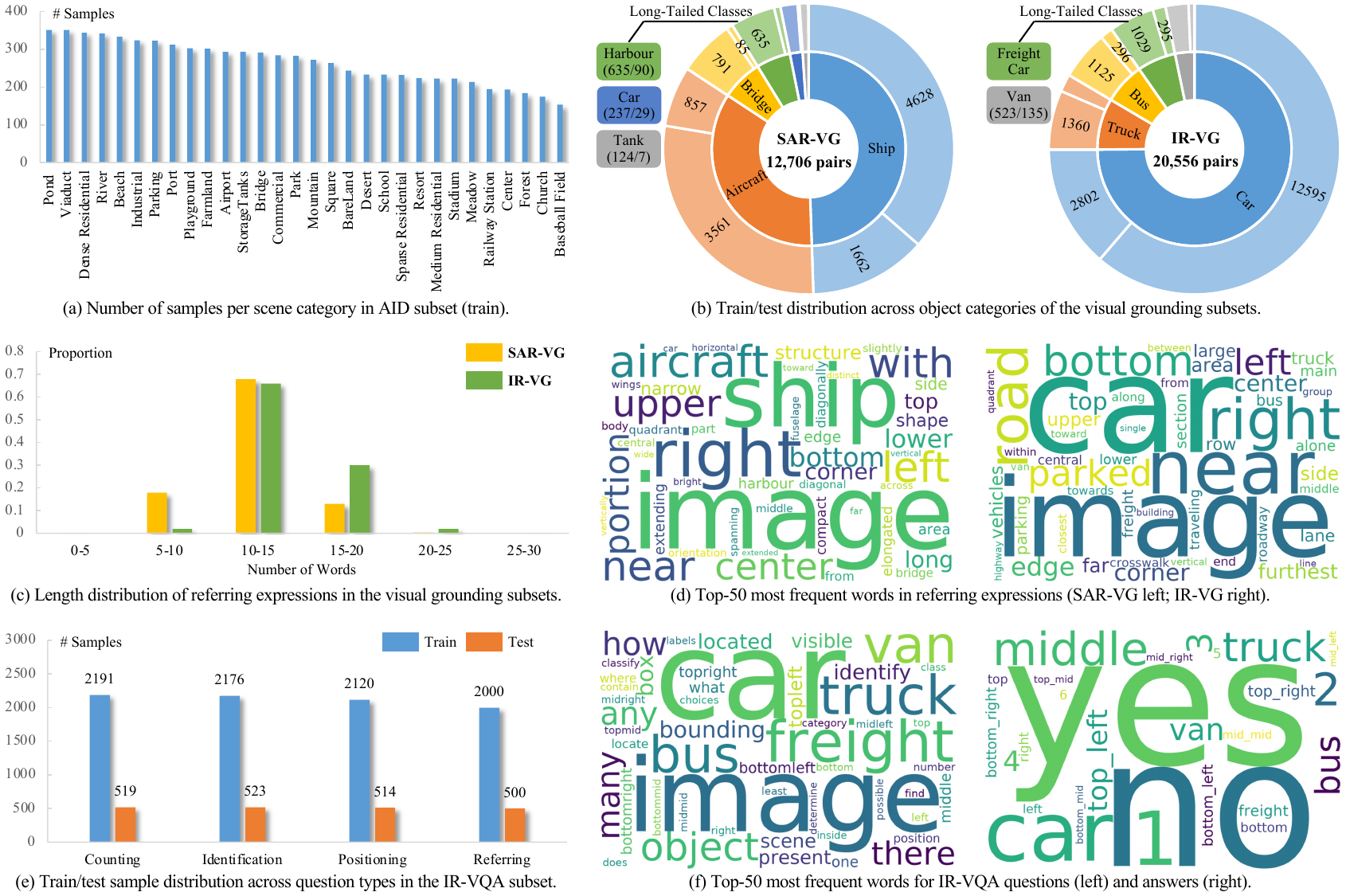}
    \caption{Statistics of the newly constructed subsets in CLeaRS. The remaining subsets are adapted from existing image-text datasets and detailed in Appendix.}
    \label{fig:clears-statistics}
\end{figure}

\textbf{Classification:} Both AID and FireRisk are formulated as classification tasks, where RS VLMs predict land use and land cover scene categories (\eg, forest, bare land, and airport) and risk levels (\eg, low, moderate, and high), respectively. The AID test split is approximately class-balanced (around 66 image-text pairs per category), while the training split shows mild imbalance, as illustrated in Fig.~\ref{fig:clears-statistics}(a). FireRisk maintains balanced per-class distributions, with 1,100 training pairs and 300 test pairs for each category.

\textbf{Visual Grounding:} The SAR-VG and IR-VG subsets target visual grounding in SAR and infrared images, respectively. Each image is paired with 1-3 referring expressions. Because SAR images contain fewer object instances per image than infrared images, comparable numbers of images ($\sim$10k) yield substantially fewer image-text pairs, \ie, 12,706 for SAR-VG versus 20,556 for IR-VG. Both subsets suffer from severe class imbalance, inherited from their source datasets and further exacerbated by the uniqueness constraint of referring expressions. Consequently, SAR-VG mainly contains references to ships and aircraft, while IR-VG is dominated by cars. Referring expressions in both subsets are generally short, with an average length of 11.32 and 13.44 words for SAR-VG and IR-VG, respectively. The distributions of expression lengths and the top-50 most frequent words are shown in Fig.~\ref{fig:clears-statistics}(c) and (d). 

\textbf{Visual Question Answering:} The IR-VQA subset is constructed for visual question answering on infrared images and comprises four question types. Each question is associated with an unambiguous answer to avoid evaluation uncertainty. Among them, object identification questions are yes/no, instance counting questions require open-ended numerical answers, while region referring and object positioning questions are presented in a multiple-choice format. As shown in Fig.~\ref{fig:clears-statistics}(e), the subset is balanced across types, with approximately 2,100 training and 510 test samples per type. Word clouds of the top-50 most frequent words from questions and answers are presented in Fig.~\ref{fig:clears-statistics}(f).

\subsection{Evaluation Protocol and Metrics}
To evaluate continual vision-language learning, we train RS VLMs on one subset at a time and evaluate them on the test sets of all previously seen subsets after completing training on each subset. This cumulative evaluation measures both the acquisition of new knowledge and the retention of previously learned capabilities, thereby assessing the stability-plasticity trade-off of the models. Based on this setting, we design three protocols in CLeaRS to systematically study the continual learning behavior of RS VLMs.

(1) \textbf{Long-Horizon Setting:} RS VLMs are trained sequentially on all subsets in the order of AID, VRS-Cap, VRS-VG, VRS-VQA, SAR-VG, SAR-VQA, IR-VG, IR-VQA, FireRisk, and RescueNet, forming a long-horizon learning scenario. This protocol progressively expands model capabilities from breadth (handling multiple tasks within a single sensing modality), to depth (multi-modal understanding), and finally to fusion (knowledge integration for specific downstream applications), thereby mimicking the developmental trajectory of RS VLMs from general-purpose perception to application-oriented specialization.

(2) \textbf{Modality-Incremental Setting:} RS VLMs are trained by progressively introducing new sensing modalities while keeping the task fixed. Specifically, the models first learn visual grounding subsets in the order of VRS-VG, SAR-VG, and IR-VG, followed by visual question answering subsets in the same modality order, resulting in six sequential stages. This protocol isolates the effect of modality shifts and evaluates the models' ability to transfer task knowledge across heterogeneous sensing modalities while retaining knowledge acquired from earlier sensing modalities.

(3) \textbf{Task-Incremental Setting:} RS VLMs are trained under a fixed optical sensing modality while new task types are progressively introduced. Specifically, the models are sequentially trained on AID, VRS-Cap, VRS-VG, and VRS-VQA, forming four stages of task expansion. Prior studies typically define "task increments" by splitting semantic categories or data distributions within a single interpretation task and rely on explicit task identifiers during inference~\cite{lu2021lil,rui2023dilrs,feng2021continual}. In contrast, our protocol introduces genuinely new task types and requires models to infer the task solely from the instruction. 

\textbf{Evaluation Metrics:} We use standard task-specific metrics for each subset (see Appendix for details) and additionally adopt continual learning measures to assess knowledge acquisition and forgetting, including Mean Fine-tune Accuracy (MFT), Mean Final Accuracy (MFN), Mean Average Accuracy (MAA), and Backward Transfer (BWT)~\cite{chen2025sefe}. MFT, MFN, and MAA measure model accuracy at different stages of learning, namely immediately after each subset is learned, after completing the full learning sequence, and averaged over all stages, while BWT measures performance changes on previously learned subsets to reflect forgetting. Detailed mathematical definitions are provided in Appendix.

\section{Experiments}
\label{sec:exp}

\subsection{Experimental Setup}
\textbf{Evaluated Models:} We benchmark both general-purpose and remote sensing-specific vision-language models on CLeaRS. For general-purpose VLMs, we evaluate Qwen2.5-VL~\cite{Qwen2.5-VL}, MiniGPT-v2~\cite{chen2023minigpt}, and LLaVA-1.5~\cite{liu2023visual}, which have inspired several adaptations to remote sensing. Domain-specific models include GeoChat~\cite{kuckreja2024geochat} and VHM~\cite{pang2025vhm}. GeoChat is a representative attempt to ground vision-language models for remote sensing, while VHM is pre-trained on large-scale remote sensing image-text pairs with rich-content captions and exhibits stronger domain-specific understanding.

\textbf{Implementation Details:} For Qwen2.5-VL and LLaVA-1.5, we adopt their 7B variants to maintain comparable model sizes with the other evaluated models. Unless otherwise specified, we follow the default configurations of each model. To accommodate high-resolution remote sensing images, we unify the input resolution to $504\times504$, with MiniGPT-v2, LLaVA-1.5, and VHM configured to support this resolution. We initialize models from their released checkpoints, using instruction-tuned versions for all models except VHM, which starts from remote sensing pre-training. During continual learning on CLeaRS, we follow GeoChat and fine-tune each model with LoRA~\cite{hu2022lora} (rank=64) applied to the language model while freezing the vision encoder. Additional training details, including the learning rate, batch size, and number of epochs, are provided in Appendix.

\begin{table*}[ht!]
    \centering
    \caption{Benchmarking VLMs on CLeaRS. \textit{Joint} corresponds to joint training on all subsets. \textit{Inst.} reports performance immediately after each subset, while \textit{Final} denotes performance after the last subset, both under sequential fine-tuning.}
    \label{tab:clears-results}
    \resizebox{\textwidth}{!}{
    \begin{tabular}{c|ccc|ccc|ccc|ccc|ccc}
    \toprule
    \multirow{2}{*}{VLMs} & \multicolumn{3}{c|}{Qwen2.5-VL~\cite{Qwen2.5-VL}} & \multicolumn{3}{c|}{MiniGPT-v2~\cite{chen2023minigpt}} & \multicolumn{3}{c|}{LLaVA-1.5~\cite{liu2023visual}} & \multicolumn{3}{c|}{GeoChat~\cite{kuckreja2024geochat}} & \multicolumn{3}{c}{VHM~\cite{pang2025vhm}} \\ 
     & Joint & Inst. & Final  & Joint & Inst. & Final & Joint & Inst. & Final & Joint & Inst. & Final & Joint & Inst. & Final \\  \midrule
     \multicolumn{16}{c}{Long-Horizon Setting} \\ \midrule
     AID        & 84.0 & 82.1 & 80.9 & 97.4 & 97.6 & 46.0 & 94.9 & 95.3 & 76.6 & 95.3 & 95.1 & 66.9 & 95.6 & 98.8 & 16.2 \\
     VRS-Cap    & 20.6 & 12.0 & 0.1  & 13.1 & 12.1 & 0.2  & 22.6 & 17.4 & 0.4  & 22.9 & 16.1 & 0.0  & 23.4 & 17.8 & 0.4 \\
     VRS-VG     & 35.4 & 10.9 & 15.2 & 25.8 & 6.2  & 0.9  & 21.0 & 8.5  & 1.8  & 33.7 & 25.2 & 12.5 & 16.3 & 6.7  & 0.2 \\
     VRS-VQA    & 66.9 & 68.7 & 66.8 & 69.1 & 70.1 & 53.1 & 71.8 & 71.3 & 59.6 & 72.6 & 72.6 & 57.9 & 72.8 & 72.4 & 43.1 \\
     SAR-VG     & 40.3 & 26.9 & 15.1 & 49.2 & 32.3 & 4.6  & 43.0 & 27.8 & 2.0  & 53.0 & 40.1 & 16.8 & 42.9 & 19.3 &  0.3 \\
     SAR-VQA    & 57.8 & 65.8 & 57.3 & 89.1 & 89.8 & 78.0 & 85.5 & 88.6 & 74.6 & 82.8 & 84.8 & 68.3 & 89.6 & 91.0 &  66.4 \\
     IR-VG      & 41.5 & 42.4 & 35.8 & 23.1 & 17.7 & 1.0  & 24.9 & 13.2 & 2.1  & 31.3 & 22.4 & 17.3 & 21.3 & 8.7  & 0.1 \\
     IR-VQA     & 52.6 & 50.7 & 51.0 & 48.2 & 50.8 & 46.5 & 54.3 & 55.2 & 53.6 & 49.9 & 53.9 & 53.6 & 56.6 & 55.4 & 48.3 \\
     FireRisk   & 22.0 & 17.8 & 16.5 & 42.4 & 43.4 & 31.2 & 40.2 & 46.4 & 45.0 & 41.2 & 39.3 & 27.0 & 48.1 & 49.2 & 37.1 \\
     RescueNet  & 49.9 & 42.7 & 42.7 & 68.5 & 64.4 & 64.4 & 65.7 & 62.7 & 62.7 & 68.5 & 61.8 & 61.8 & 70.9 & 65.3 & 65.3 \\ \midrule
     MFT & \multicolumn{3}{c|}{44.3} & \multicolumn{3}{c|}{50.8} & \multicolumn{3}{c|}{52.0} & \multicolumn{3}{c|}{\textbf{54.4}} & \multicolumn{3}{c}{51.9} \\
     MFN &\multicolumn{3}{c|}{\textbf{38.2}} & \multicolumn{3}{c|}{32.6} & \multicolumn{3}{c|}{37.9} & \multicolumn{3}{c|}{\textbf{38.2}} & \multicolumn{3}{c}{27.8} \\
     MAA & \multicolumn{3}{c|}{46.4} & \multicolumn{3}{c|}{46.4} & \multicolumn{3}{c|}{46.5} & \multicolumn{3}{c|}{\textbf{49.6}} & \multicolumn{3}{c}{35.3} \\
     BWT & \multicolumn{3}{c|}{\textbf{-6.9}} & \multicolumn{3}{c|}{-20.2} & \multicolumn{3}{c|}{-15.7} & \multicolumn{3}{c|}{-18.0} & \multicolumn{3}{c}{-26.8} \\ \midrule
     \multicolumn{16}{c}{Modality-Incremental Setting} \\ \midrule
     VRS-VG     & 33.1 & 10.5 & 18.3 & 24.0 & 6.4  & 0.6  & 20.0 & 9.5  & 2.3  & 31.6 & 22.6 & 6.7  & 17.6 & 5.7 & 0.7 \\
     SAR-VG     & 43.5 & 27.6 & 22.1 & 45.9 & 33.6 & 3.7  & 42.6 & 19.8 & 3.1  & 44.5 & 41.6 & 13.2 & 40.8 & 12.7 & 1.2 \\
     IR-VG      & 46.7 & 40.7 & 38.8 & 21.9 & 20.8 & 0.9  & 24.1 & 17.0 & 1.9  & 26.9 & 23.4 & 2.9  & 20.7 & 12.6 & 0.3 \\
     VRS-VQA    & 66.5 & 65.7 & 59.8 & 68.7 & 68.9 & 57.7 & 70.6 & 69.4 & 58.8 & 70.8 & 71.8 & 63.0 & 71.7 & 72.1 & 44.3 \\
     SAR-VQA    & 55.8 & 65.9 & 40.4 & 87.6 & 89.6 & 80.2 & 85.8 & 89.4 & 81.1 & 79.1 & 85.9 & 77.4 & 89.3 & 91.9 & 42.5 \\
     IR-VQA     & 51.6 & 50.4 & 50.4 & 48.0 & 51.6 & 51.6 & 53.5 & 53.6 & 53.6 & 50.0 & 53.6 & 53.6 & 54.2 & 56.0 & 56.0 \\ \midrule
     MFT & \multicolumn{3}{c|}{43.5} & \multicolumn{3}{c|}{45.2} & \multicolumn{3}{c|}{43.1} & \multicolumn{3}{c|}{\textbf{49.8}} & \multicolumn{3}{c}{41.8} \\
     MFN & \multicolumn{3}{c|}{\textbf{38.3}} & \multicolumn{3}{c|}{32.5} & \multicolumn{3}{c|}{33.5} & \multicolumn{3}{c|}{36.1} & \multicolumn{3}{c}{24.2} \\
     MAA & \multicolumn{3}{c|}{29.1} & \multicolumn{3}{c|}{22.2} & \multicolumn{3}{c|}{20.0} & \multicolumn{3}{c|}{\textbf{29.3}} & \multicolumn{3}{c}{15.7} \\
     BWT & \multicolumn{3}{c|}{\textbf{-6.2}} & \multicolumn{3}{c|}{-15.2} & \multicolumn{3}{c|}{-11.6} & \multicolumn{3}{c|}{-16.4} & \multicolumn{3}{c}{-21.2} \\ \midrule
     \multicolumn{16}{c}{Task-Incremental Setting} \\ \midrule
     AID        & 83.2 & 82.1 & 81.2 & 96.6 & 97.6 & 82.1 & 94.7 & 95.3 & 74.4 & 94.9 & 95.1 & 80.8 & 96.0 & 98.8 & 67.9 \\
     VRS-Cap    & 20.3 & 12.0 & 0.0  & 14.0 & 12.1 & 7.6  & 21.9 & 17.4 & 0.0  & 22.2 & 16.1 & 0.1  & 22.5 & 17.8 & 0.0 \\
     VRS-VG     & 25.5 & 10.9 & 10.3 & 11.7 & 6.2  & 2.2  & 14.1 & 8.5  & 7.0  & 27.9 & 25.2 & 17.1 & 8.5 & 6.7 & 0.6 \\
     VRS-VQA    & 67.2 & 68.7 & 68.7 & 68.7 & 70.1 & 70.1 & 71.3 & 71.3 & 71.3 & 72.3 & 72.6 & 72.6 & 71.8 & 72.4 & 72.4 \\ \midrule     
     MFT & \multicolumn{3}{c|}{49.2} & \multicolumn{3}{c|}{52.4} & \multicolumn{3}{c|}{56.6} & \multicolumn{3}{c|}{\textbf{60.4}} &  \multicolumn{3}{c}{57.6} \\
     MFN & \multicolumn{3}{c|}{40.0} & \multicolumn{3}{c|}{\textbf{44.2}} & \multicolumn{3}{c|}{38.2} & \multicolumn{3}{c|}{42.7} &  \multicolumn{3}{c}{35.3} \\
     MAA & \multicolumn{3}{c|}{55.6} & \multicolumn{3}{c|}{59.9} & \multicolumn{3}{c|}{61.0} & \multicolumn{3}{c|}{\textbf{64.2}} &  \multicolumn{3}{c}{45.0} \\
     BWT & \multicolumn{3}{c|}{-12.3} & \multicolumn{3}{c|}{\textbf{-10.9}} & \multicolumn{3}{c|}{-24.5} & \multicolumn{3}{c|}{-23.6} &  \multicolumn{3}{c}{-29.8} \\
     \bottomrule
    \end{tabular}
    }
\end{table*}

\subsection{How Do Existing VLMs Perform in CLeaRS?}
To comprehensively examine the continual learning capability of VLMs in remote sensing scenarios, we evaluate both generic and domain-specific VLMs under the three evaluation protocols defined in CLeaRS. The quantitative results are reported in Table~\ref{tab:clears-results}. Based on these results, we make the following observation:

\begin{figure}[t]
    \centering
    \includegraphics[width=\textwidth]{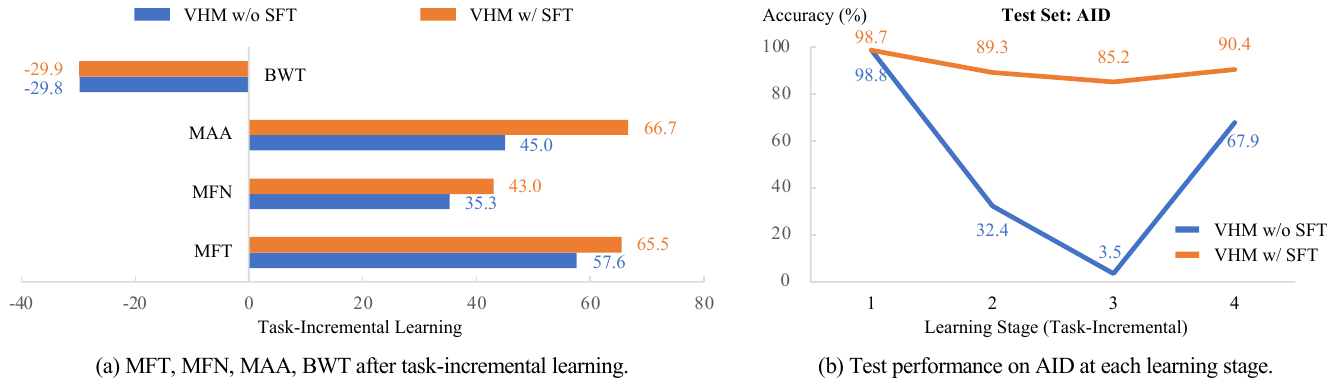}
    \caption{Comparison of VHM with and without supervised fine-tuning in the task-incremental setting.}
    \label{fig:vhm-comparison}
\end{figure}

\textbf{First}, all evaluated VLMs obtain negative BWT scores across the three settings, indicating consistent forgetting during continual learning on CLeaRS. Moreover, VHM exhibits the largest absolute BWT values under all settings, suggesting the most severe forgetting. In contrast, Qwen2.5-VL and MiniGPT-v2 show relatively milder forgetting under specific settings. One possible explanation for the pronounced forgetting observed in VHM is that its frozen weights have only undergone pre-training, whereas the frozen components of the other models have further benefited from supervised fine-tuning. To verify this conjecture, we additionally compare VHM with and without supervised fine-tuning (SFT) under the task-incremental setting (Fig.~\ref{fig:vhm-comparison}). The results show that both variants suffer from similar levels of forgetting. However, VHM with SFT achieves substantially higher final performance than the non-fine-tuned variant, even surpassing the other three models. These findings indicate that our initial explanation does not hold: the severe forgetting of VHM is not solely due to the lack of supervised fine-tuning. 

\textbf{Second}, the severity of forgetting varies across evaluation protocols. Interestingly, forgetting does not monotonically increase with the length of the learning sequence. For four of the five models, the most severe forgetting occurs in the shortest setting, \ie, the task-incremental setting. In contrast, MiniGPT-v2 exhibits the largest forgetting under the long-horizon setting, which involves the largest number of sequential learning stages. This observation suggests that forgetting in VLMs is not solely determined by the sequence length, but may also be influenced by task boundary transitions introduced in different protocols. The inconsistent trends across models further indicate that model-specific design and training factors may influence their stability (\ie, their ability to retain previously acquired knowledge) under different sequential learning protocols.

\textbf{Third}, we further analyze model plasticity, namely, how effectively a model acquires new knowledge during sequential learning. To this end, we compare the average accuracy after joint training with the MFT score, which averages performance on each subset immediately after it is learned. Across all settings, joint training consistently achieves higher average accuracy than MFT for all models. Nevertheless, for every model, there are subsets on which individually fine-tuned models outperform joint training on the same subsets. For example, on SAR-VQA under the long-horizon setting, all five models achieve higher accuracy with independent fine-tuning than with joint training, with Qwen2.5-VL exhibiting the largest margin (65.8 vs. 57.8). In remote sensing continual learning research, joint training is commonly used as an upper-bound reference~\cite{sun2025mitigating,shan2022class,xie2024missnet}. However, our findings suggest that this convention may not directly transfer to VLM-based continual learning: although joint training improves overall performance, it can dilute subset-specific optimization.

\textbf{Finally}, we evaluate the continual learning capability of different models on our benchmark using MFN and MAA. MFN reflects the average accuracy after completing the full learning sequence, while MAA averages model performance across all learning stages, reflecting its behavior throughout the learning process. Under the long-horizon, modality-incremental, and task-incremental settings, the highest MFN is achieved by GeoChat, Qwen2.5-VL, and MiniGPT-v2, respectively. Notably, the model with the highest MFN does not necessarily attain the highest MAA. For example, under the task-incremental setting, LLaVA-1.5 achieves a high MAA (61.0), yet its MFN (38.2) is substantially lower than that of MiniGPT-v2, which achieves the highest MFN (44.2). This discrepancy indicates that strong final performance does not necessarily imply consistently strong performance throughout the learning process. A model may maintain stable accuracy across intermediate stages (reflected by high MAA) yet exhibit weaker retention at the end of the sequence, or vice versa.

\subsection{What Makes Continual Learning Challenging for RS VLMs?}
To gain insights into forgetting in RS VLMs, we conduct controlled experiments to investigate the effects of task heterogeneity, instruction variation, and modality transitions. We choose VHM as the representative model. While GeoChat achieves stronger overall performance, its supervised fine-tuning data partially overlaps with our benchmark in terms of imagery, which may confound the interpretation of forgetting. In contrast, VHM's pre-training data has no confirmed overlap with CLeaRS, providing a cleaner starting point.

\begin{figure}[t]
    \centering
    \includegraphics[width=\textwidth]{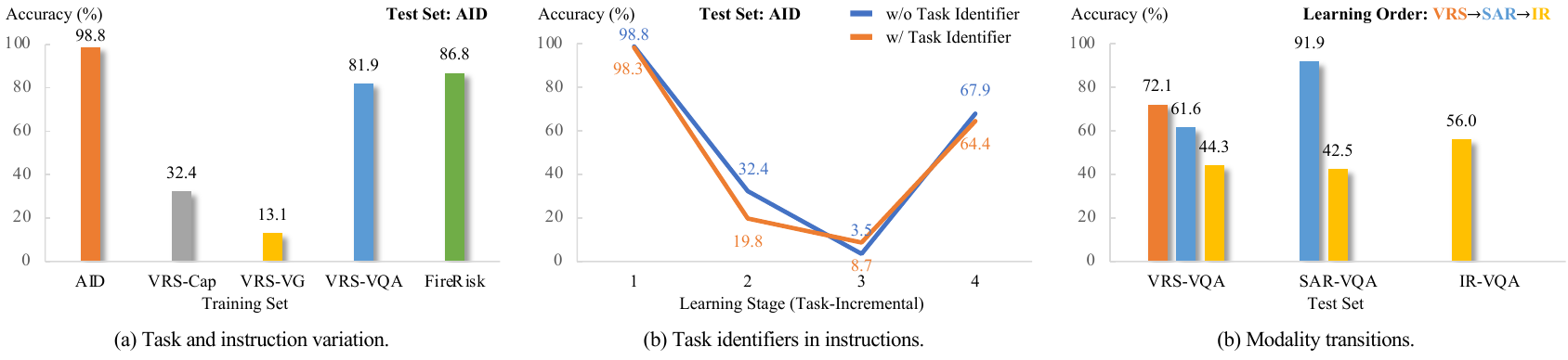}
    \caption{Analysis of factors contributing to forgetting in RS VLMs.}
    \label{fig:forgetting-factors}
\end{figure}

Fig.~\ref{fig:forgetting-factors}(a) reports the AID performance after fine-tuning on each of VRS-Cap, VRS-VG, and VRS-VQA. Training on VRS-VQA results in the smallest performance drop on AID, whereas VRS-VG causes the most severe forgetting. Task differences are accompanied by variations in instruction formats. For example, AID requires predicting a single category, while VRS-VG involves bounding box coordinates. Prior work~\cite{chen2024coin} suggests that forgetting mainly arises from instruction misalignment rather than true knowledge loss. To examine this, we further perform continual learning on FireRisk, which adopts instruction templates similar to AID. Despite the aligned format, performance still drops by more than 10\%, although retention is better than training on VRS-VQA. This indicates that instruction similarity alone is insufficient to prevent forgetting. Moreover, intrinsic task differences preclude enforcing identical instruction formats.

Motivated by this observation, we incorporate explicit task identifiers into the instructions of AID, VRS-Cap, VRS-VG, and VRS-VQA and perform four-stage sequential fine-tuning. As shown in Fig.~\ref{fig:forgetting-factors}(b), although task identifiers improve overall performance under joint training~\cite{pang2025vhm,li2024vrsbench}, they provide limited benefit for mitigating forgetting in continual learning, and sometimes even result in large performance drops. Finally, we analyze sensing modality transitions. As illustrated in Fig.~\ref{fig:forgetting-factors}(c), even under the same task and similar instruction formats, modality differences still lead to pronounced forgetting of previously learned visual concepts. These results show that forgetting in RS VLMs cannot be attributed to a single factor, but is jointly affected by task heterogeneity, instruction variation, and modality transitions. 

\subsection{Are Continual Learning Methods for VLMs Effective in RS?}
Results in Table~\ref{tab:clears-results} show that RS VLMs suffer from catastrophic forgetting on CLeaRS. Given that most RS VLMs rely on parameter-efficient fine-tuning for downstream adaptation~\cite{kuckreja2024geochat,zhan2025skyeyegpt,muhtar2024lhrs,luo2024skysensegpt}, we further evaluate representative parameter-efficient continual learning methods using VHM as the backbone model. Detailed hyperparameter settings are provided in Appendix.

\textbf{Benchmarking Continual Learning Methods on CLeaRS:} Specifically, we benchmark MoELoRA~\cite{chen2024coin}, HiDe-LLaVA~\cite{guo2025hide}, and SEFE~\cite{chen2025sefe}. As shown in Table~\ref{tab:clears-cl-method}, all methods exhibit negative BWT scores across the three settings, indicating that catastrophic forgetting in remote sensing continual learning remains unresolved. In the modality-incremental setting, the methods achieve noticeable improvements over sequential fine-tuning, improving the BWT from -21.2 to approximately -12.2 and increasing the MFN from 24.2 to around 32.0. However, in the long-horizon and task-incremental settings, their mitigation effect becomes marginal. In several cases, they even lead to more severe forgetting and lower overall performance than unconstrained sequential fine-tuning. These results suggest that existing continual learning methods lack stable performance across different remote sensing continual learning scenarios.

Regarding method-wise comparison, no single method consistently outperforms others across all settings. SEFE generally achieves the highest MFN, yet its BWT remains close to sequential fine-tuning in the long-horizon and task-incremental settings, indicating limited improvement in forgetting mitigation. HiDe-LLaVA mitigates forgetting more effectively in the long-horizon and modality-incremental settings, but does not consistently achieve strong overall performance. MoELoRA behaves similarly to other methods, slightly improving forward knowledge acquisition in the long-horizon setting. The results suggest that effectively balancing knowledge retention and adaptation remains a central challenge for continual learning in remote sensing.

\textbf{Learning Order Sensitivity of Continual Learning Methods:} Beyond the performance variation observed across different settings, we further analyze the sensitivity of continual learning methods to learning order. In the long-horizon setting, which contains the longest learning sequence, we permute the subset order to construct multiple learning sequences (see Appendix) and assess performance consistency across different orderings. As shown in Table~\ref{tab:clears-learning-order}, varying the learning order introduces moderate performance fluctuations across all methods. Nevertheless, BWT remains substantially negative under all sequences, even for the best-performing ordering of each method. This indicates that subset reordering alone cannot fundamentally alleviate catastrophic forgetting. 

\begin{table*}[t]
    \centering
    \caption{Performance of continual learning methods on CLeaRS. All methods use VHM as the backbone model for fair comparison.}
    \label{tab:clears-cl-method}
    \resizebox{\textwidth}{!}{
    \begin{tabular}{c|cccc|cccc|cccc}
    \toprule
    \multirow{2}{*}{Methods} & \multicolumn{4}{c|}{Long-Horizon} & \multicolumn{4}{c|}{Modality-Incremental} & \multicolumn{4}{c}{Task-Incremental} \\ 
     & MFT & MFN & MAA & BWT & MFT & MFN & MAA & BWT & MFT & MFN & MAA & BWT \\ \midrule
    Sequential FT                 & 51.9 & 27.8 & 35.3 & -26.8 & 41.8 & 24.2 & 15.7 & -21.2 & 57.6 & 35.3 & 45.0 & -29.8 \\
    MoELoRA~\cite{chen2024coin}   & \textbf{53.2} & \textbf{29.1} & \textbf{37.2} & -26.8 & 41.8 & 31.6 & 17.1 & -12.2 & 56.7 & 34.7 & 46.1 & -29.4 \\
    HiDe-LLaVA~\cite{guo2025hide} & 49.4 & 27.7 & 32.8 & \textbf{-24.1} & 40.9 & 31.2 & 16.0 & \textbf{-11.6} & 56.1 & 31.4 & 43.9 & -32.9 \\
    SEFE~\cite{chen2025sefe}      & 52.9 & 27.5 & \textbf{37.2} & -28.3 & \textbf{43.8} & \textbf{33.1} & \textbf{18.2} & -12.8 & \textbf{59.1} & \textbf{37.2} & \textbf{49.7} & \textbf{-29.1} \\
    \bottomrule
    \end{tabular}
    }
\end{table*}

\begin{table*}[t]
    \centering
    \caption{Performance of continual learning methods under different learning orders on CLeaRS (long-horizon). Numbered markers denote learning sequences (see Appendix).}
    \label{tab:clears-learning-order}
    \resizebox{\textwidth}{!}{
    \begin{tabular}{c|cccc|cccc|cccc}
    \toprule
    \multirow{2}{*}{Methods} & \multicolumn{4}{c|}{MoELoRA~\cite{chen2024coin}} & \multicolumn{4}{c|}{HiDe-LLaVA~\cite{guo2025hide}} & \multicolumn{4}{c}{SEFE~\cite{chen2025sefe}} \\
     & MFT & MFN & MAA & BWT & MFT & MFN & MAA & BWT & MFT & MFN & MAA & BWT \\ \midrule
    \hollowcircled{1} & 53.2 & 29.1 & 37.2 & -26.8 & 49.4 & \textbf{27.7} & \textbf{32.8} & \textbf{-24.1} & 52.9 & 27.5 & 37.2 & -28.3  \\    
    \hollowcircled{2} & 53.1 & \textbf{32.3} & 36.6 & \textbf{-23.1} & \textbf{50.1} & 26.1 & \textbf{32.8} & -26.7 & 53.1 & 28.0 & 37.2 & -27.9  \\   
    \hollowcircled{3} & \textbf{53.5} & 31.5 & \textbf{37.8} & -24.4 & 47.7 & 22.4 & 27.7 & -28.1 & \textbf{53.6} & \textbf{30.5} & \textbf{37.6} & \textbf{-25.6}  \\   \midrule            
    \makecell{Mean \\ $\pm$Std}      & \makecell{53.3 \\$\pm$0.2 } & \makecell{31.0 \\ $\pm$1.4} & \makecell{37.2 \\ $\pm$0.5} & \makecell{-24.8 \\ $\pm$1.5} & \makecell{49.1 \\ $\pm$1.0} & \makecell{25.4 \\ $\pm$2.2} & \makecell{31.1 \\ $\pm$2.4} & \makecell{-26.3 \\ $\pm$1.7} & \makecell{53.2 \\ $\pm$0.3} & \makecell{28.7 \\ $\pm$1.3} & \makecell{37.3 \\ $\pm$0.2} & \makecell{-27.3 \\ $\pm$1.2}  \\
    \bottomrule
    \end{tabular}
    }
\end{table*}

\section{Conclusion}
In this paper, we introduce CLeaRS, the first benchmark for systematically investigating continual learning in RS VLMs. CLeaRS consists of ten subsets covering diverse interpretation tasks, sensing modalities, and application scenarios. We define three evaluation protocols, including long-horizon, modality-incremental, and task-incremental settings, and benchmark multiple vision-language models and representative continual learning methods. The results demonstrate that RS VLMs suffer from catastrophic forgetting in continual learning scenarios. Existing continual learning methods, when adapted to RS VLMs, struggle to balance stability and plasticity and fail to deliver consistent gains over sequential fine-tuning across different settings and learning orders. We hope our CLeaRS will encourage the development of continual learning methods tailored to RS VLMs.

\newpage

{
\small
\bibliographystyle{unsrt}
\bibliography{ref}
}

\clearpage
\setcounter{page}{1}
\appendix

\section*{Contents}
\noindent
A. \hspace{0.2cm} Benchmark Curation Details \dotfill \pageref{benchmark-curation-details} \\
B. \hspace{0.2cm} Additional Benchmark Details \dotfill \pageref{additional-benchmark-details} \\
C. \hspace{0.2cm} Experimental Details \dotfill \pageref{experiment-details} \\
D. \hspace{0.2cm} Additional Experimental Results and Analysis \dotfill \pageref{additional-experimental-results} \\

\section{Benchmark Curation Details}
\label{benchmark-curation-details}

\subsection{Source Datasets}
To ensure the accessibility of CLeaRS, we construct the benchmark using several publicly available remote sensing datasets, including AID~\cite{xia2017aid}, VRSBench~\cite{li2024vrsbench}, SARDet-100K~\cite{li2024sardet}, SARLANG-1M~\cite{wei2025sarlang}, DroneVehicle~\cite{sun2022drone}, FireRisk~\cite{shen2023firerisk}, and RescueNet-VQA~\cite{sarkar2023rescuenet}. Table~\ref{tab:source-dataset} summarizes their statistics. These datasets span diverse geographic regions, sensing modalities, and spatial resolutions, enabling CLeaRS to better simulate the evolving remote sensing data streams encountered during model deployment.

\subsection{Prompt Design for Referring Expressions}
\begin{figure}[ht!]
    \centering
    \includegraphics[width=\textwidth]{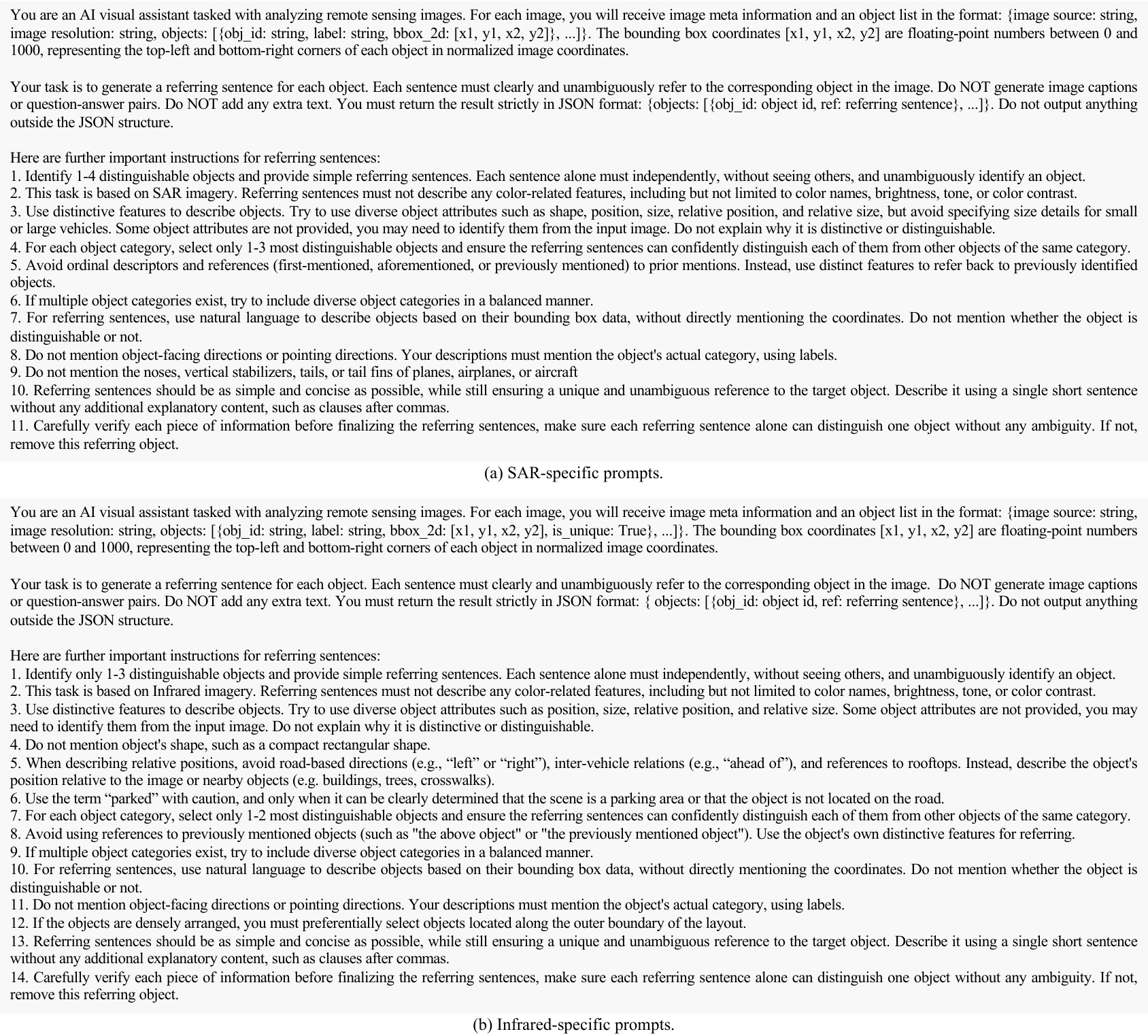}
    \caption{Prompts for referring expression generation in SAR and infrared imagery.}
    \label{fig:prompt-vg}
\end{figure}

Based on SARDet-100K and DroneVehicle, we construct visual grounding subsets for SAR and infrared imagery. Inspired by the prompt design in \cite{li2024vrsbench}, we develop modality-specific prompts to guide state-of-the-art vision-language models (Qwen3-VL-Plus~\cite{Qwen3-VL} and Gemini-3-Flash~\cite{gemini3}) to generate referring expressions (see Fig.~\ref{fig:prompt-vg}). To ensure unambiguous grounding, we use GLM-4.6V~\cite{hong2025glm} to infer the target object from each generated expression and automatically filter out ambiguous cases, followed by manual verification. For consistency, the absolute bounding box coordinates in all three visual grounding subsets are normalized to the range [0,1000].

\subsection{Instruction Templates for Visual Question Answering}
To support visual question answering on infrared imagery, we create IR-VQA with four question types: object identification, instance counting, object positioning, and region referring. Table~\ref{tab:instruction-template} shows the instruction templates for each task. For object identification, we ensure a balanced distribution between positive and negative samples (1:1). For instance counting, images containing more than seven instances are excluded. For object positioning, we select images where the target category appears only once and determine the positional label using several spatial partitions, including cross, 3$\times$3 grid, horizontal tripartition, and vertical tripartition.

\begin{table}[ht]
    \centering
    \caption{Statistics of the source datasets used to construct CLeaRS.}
    \label{tab:source-dataset}
    \resizebox{\textwidth}{!}{
    \begin{tabular}{ccccc}
    \toprule
    Dataset & Task & \#Samples & Resolution (m) & License \\ \midrule
    AID~\cite{xia2017aid} & Scene Classification & 10,000 & 0.5$\sim$8 & - \\
    VRSBench~\cite{li2024vrsbench} & \makecell{Image Captioning \\ Visual Grounding \\ Visual Question Answering} & 205,307 & 0.1$\sim$30 & CC BY 4.0 \\
    SARDet-100K~\cite{li2024sardet} & Object Detection & 116,598 & 0.1$\sim$25 & CC BY-NC 4.0 \\
    SARLANG-1M~\cite{wei2025sarlang} & \makecell{Image Captioning \\ Visual Question Answering} & 118,331 & 0.1$\sim$25 & - \\
    DroneVehicle~\cite{sun2022drone} & Vehicle Detection & 56,878 & -  & - \\
    FireRisk~\cite{shen2023firerisk} & Fire Risk Assessment & 91,872 & - & - \\
    RescueNet-VQA~\cite{sarkar2023rescuenet} & Damage Assessment & 103,192 & 0.015 & CC BY-NC-ND 4.0 \\
    \bottomrule
    \end{tabular}
    }
\end{table}

\begin{table}[t]
    \centering
    \caption{Instruction templates and example answers in IR-VQA.}
    \label{tab:instruction-template}
    \resizebox{\textwidth}{!}{
    \begin{tabular}{ccc}
    \toprule
    Question Type & Instruction Template & Example Answers \\ \midrule
    Object Identification & Is there at least one <specific category> present in the scene? & "Yes" or "No"\\
    Instance Counting & How many <specific category> can you find in this image? & Number \\
    Object Positioning & Where is the <specific category> located in this image? & "Left", "Middle", "Right" \\
    Region Referring & Classify the object located at the bounding box <detailed coordinates>? & \makecell{"Car", "Bus", "Van",\\ "Truck", "Freight Car"} \\
    \bottomrule
    \end{tabular}
    }
\end{table}

\section{Additional Benchmark Details}
\label{additional-benchmark-details}

\subsection{Subset Statistics}
We further present the statistics of five additional subsets in CLeaRS: VRS-Cap, VRS-VQA, VRS-VG, SAR-VQA, and RescueNet, which are sampled from existing remote sensing vision-language datasets.

\begin{figure}[t!]
    \centering
    \includegraphics[width=\textwidth]{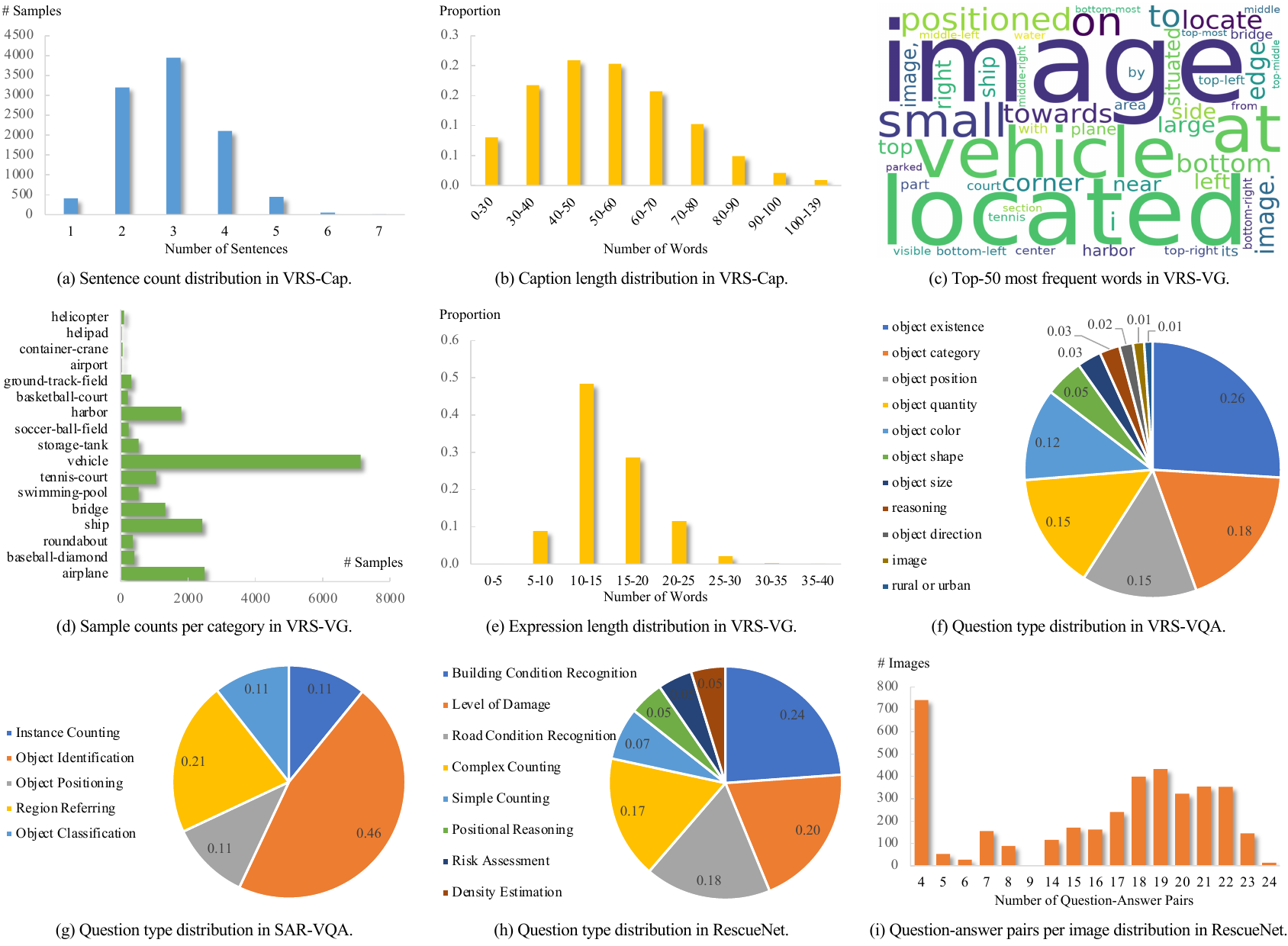}
    \caption{Statistics of additional subsets in CLeaRS.}
    \label{fig:subset-statistics}
\end{figure}

\textbf{Image Captioning:} In VRS-Cap, captions contain 2.92 sentences on average with an average length of 52.92 words. Fig.~\ref{fig:subset-statistics}(a-b) shows the distributions of sentence count and caption lengths.

\textbf{Visual Grounding:} The object referring expressions in VRS-VG leverage diverse object attributes to ensure unique reference. The subset covers 17 object categories, with an average expression length of 14.33 words. Fig.~\ref{fig:subset-statistics}(c-e) shows the top-50 most frequent words, the sample counts for each category, and the distribution of expression lengths.

\textbf{Visual Question Answering:} VRS-VQA, SAR-VQA, and RescueNet are three subsets designed for visual question answering. In VRS-VQA, most optical images are associated with 2-5 question-answer pairs, covering diverse object-related attributes. Fig.~\ref{fig:subset-statistics}(f) shows the proportion of different question types. In SAR-VQA, most SAR images contain 1-4 question-answer pairs. Due to the limited visible attributes in SAR images, the question focuses on a smaller set of object-level reasoning tasks. Fig.~\ref{fig:subset-statistics}(g) shows the proportion of each question type. In RescueNet, eight question types are designed for optical images collected after hurricanes. Fig.~\ref{fig:subset-statistics}(h-i) illustrates the distribution of the number of question-answer pairs per image, and the proportion of each question type.

\begin{figure}[t!]
    \centering
    \includegraphics[width=\textwidth]{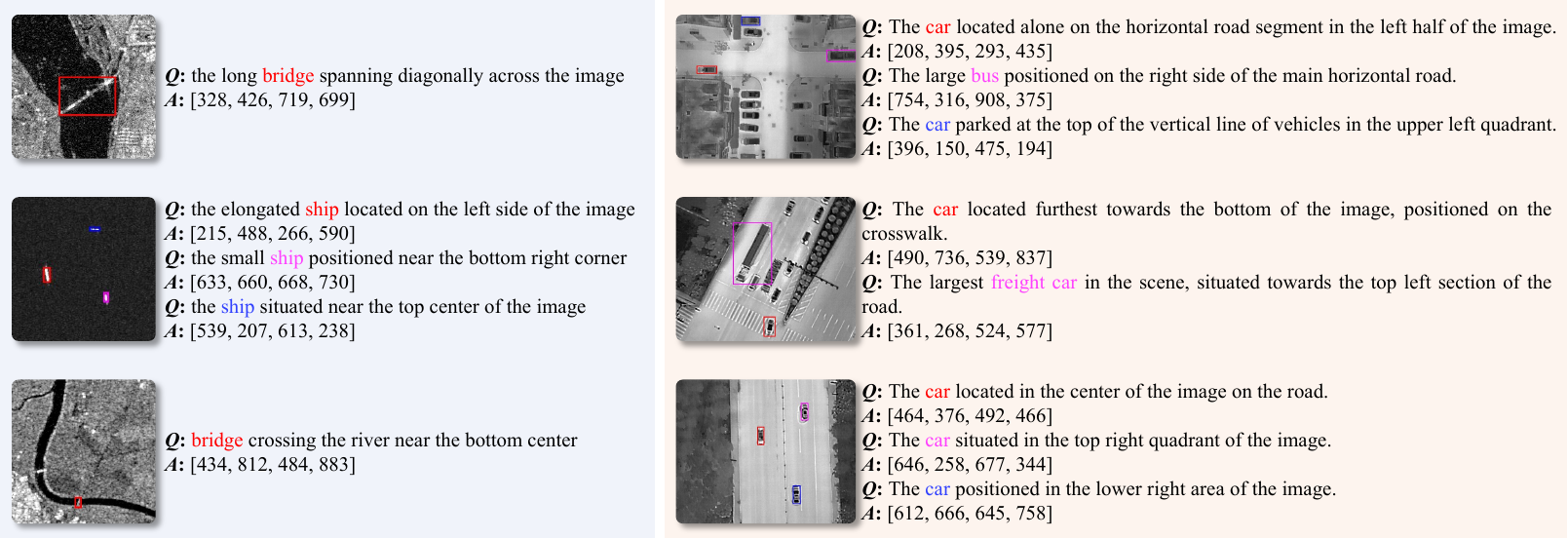}
    \caption{Examples from the SAR-VG (left) and IR-VG (right) subsets in CLeaRS.}
    \label{fig:example-vg}
\end{figure}

\subsection{Additional Examples of SAR-VG and IR-VG Subsets}
Public visual grounding datasets for SAR and infrared images remain limited. As part of CLeaRS, we construct SAR-VG and IR-VG subsets. Fig.~\ref{fig:example-vg} presents additional examples from these subsets.

\section{Experimental Details}
\label{experiment-details}
\subsection{Training Details}
All experiments are conducted on four NVIDIA H20 GPUs (96GB). For the five evaluated models, the learning rate is set to 2e-5, except for MiniGPT-v2~\cite{chen2023minigpt}, which uses an initial learning rate of 2e-4 with a minimum learning rate of 2e-5. These configurations are kept consistent across all sequential and joint learning. For joint learning, models are trained for 3 epochs. For sequential learning, the number of epochs is 3 for all subsets except RescueNet, where the model is trained for 1 epoch. The batch size is set to 128.

For continual learning methods based on VHM~\cite{pang2025vhm}, the learning rate, batch size, and number of epochs follow the same configurations as the unconstrained VHM. For MoELoRA~\cite{chen2024coin} and HiDe-LLaVA~\cite{guo2025hide}, the number of experts varies with the number of learning stages in the evaluation protocols. Specifically, the numbers of experts are 10, 6, and 4 for the long-horizon, modality-incremental, and task-incremental settings, respectively, with the rank set to 16. For SEFE~\cite{chen2025sefe}, the rank is set to 64 for all settings.

\subsection{Evaluation Metrics}
For classification (AID and FireRisk) and visual question answering (VRS-VQA, SAR-VQA, IR-VQA, and RescueNet) tasks, we report accuracy. For visual grounding tasks (VRS-VG, SAR-VG, and IR-VG), we adopt accuracy@0.5, where a prediction is considered correct if the intersection over union between the predicted and ground-truth bounding boxes exceeds 0.5. For the image captioning task (VRS-Cap), we report CIDEr, which is widely used in captioning benchmarks due to its stronger correlation with human judgment and wider dynamic range. However, since the scale of CIDEr differs from accuracy, it cannot be directly used to compute continual learning metrics (MFT, MFN, MAA, and BWT). To make the metrics comparable across tasks, we normalize the captioning scores using the highest CIDEr value (33.9) reported on the original VRSBench-Cap~\cite{li2024vrsbench} dataset as the upper bound. The normalized scores are then used when computing the continual learning metrics. This normalization aims to evaluate a model's ability to retain captioning knowledge during continual learning, rather than its absolute captioning performance.

Let $A_i^j$ denote the performance of the model after learning stage $j$ on subset $i$, where $i\leq j$, and let $T>1$ be the total number of learning stages. The four continual learning metrics are computed as follows:

\begin{align}
    \mathrm{MFT} &= \frac{1}{T}\sum_{i=1}^T A_i^i, \\
    \mathrm{MFN} &= \frac{1}{T}\sum_{i=1}^T A_i^T, \\
    \mathrm{MAA} &= \frac{1}{T}\sum_{j=1}^T \frac{1}{j} \sum_{i=1}^j A_i^j, \\
    \mathrm{BWT} &= \frac{1}{T-1} \sum_{i=1}^{T-1} (A_i^T-A_i^i).
\end{align}

\subsection{Learning Orders in the Long-Horizon Setting}
Table~\ref{tab:learning-order-setup} lists the learning orders used in the long-horizon setting to evaluate the sensitivity of continual learning methods to learning order.

\section{Additional Experimental Results and Analysis}
\label{additional-experimental-results}

\subsection{Zero-Shot Evaluation to Check Data Leakage}
Since CLeaRS is constructed from publicly available datasets, we conduct a zero-shot evaluation to examine potential data leakage between the training data of selected VLMs and our benchmark. Table~\ref{tab:result-zero-shot} reports the zero-shot performance on CLeaRS. Among general-domain VLMs, Qwen2.5-VL~\cite{Qwen2.5-VL} achieves strong performance on AID and VRS-VQA (both above 60\% accuracy), while LLaVA-1.5 performs reasonably well on VRS-VQA (57.3\%). Notably, the source datasets (AID~\cite{xia2017aid} and DIOR~\cite{li2020object}) of these subsets are widely used in the remote sensing community. For RS VLMs, the supervised fine-tuning data of GeoChat~\cite{kuckreja2024geochat} overlaps with VRS-Cap, VRS-VG, and VRS-VQA at the image level. However, due to differences in textual annotations, GeoChat shows relatively good performance only on VRS-VQA (54.1\%). In contrast, VHM has not undergone supervised fine-tuning, and its pretraining data does not overlap with these subsets, resulting in very poor zero-shot performance across the benchmark. Overall, the selected VLMs achieve relatively good performance on only a few subsets while performing poorly on most others. This observation suggests that CLeaRS introduces limited data leakage and serves as a suitable testbed for investigating continual vision-language learning in remote sensing.

 \begin{table}[t]
    \centering
    \caption{Different learning orders in CLeaRS (long-horizon).}
    \label{tab:learning-order-setup}
    \begin{tabular}{c|m{10cm}}
    \toprule
    ID & Learning Order \\ \midrule
    \hollowcircled{1} & AID$\rightarrow$VRS-Cap$\rightarrow$VRS-VG$\rightarrow$VRS-VQA$\rightarrow$SAR-VG$\rightarrow$SAR-VQA$\rightarrow$IR-VG$\rightarrow$IR-VQA$\rightarrow$FireRisk$\rightarrow$RescueNet \\
    \hollowcircled{2} & VRS-VQA$\rightarrow$AID$\rightarrow$VRS-VG$\rightarrow$VRS-Cap$\rightarrow$IR-VG$\rightarrow$SAR-VG$\rightarrow$SAR-VQA$\rightarrow$IR-VQA$\rightarrow$RescueNet$\rightarrow$FireRisk \\
    \hollowcircled{3} & VRS-Cap$\rightarrow$VRS-VQA$\rightarrow$AID$\rightarrow$VRS-VG$\rightarrow$IR-VG$\rightarrow$IR-VQA$\rightarrow$SAR-VQA$\rightarrow$SAR-VG$\rightarrow$FireRisk$\rightarrow$RescueNet \\
    \bottomrule
    \end{tabular}
\end{table}

\begin{table}[t]
    \centering
    \caption{Zero-Shot Performance of VLMs on the CLeaRS benchmark. VHM w/o SFT denotes the model without supervised fine-tuning. Bounding box coordinates in VRS-VG, SAR-VG, and IR-VG are converted to match the formats used by each model.}
    \label{tab:result-zero-shot}
    \resizebox{\textwidth}{!}{
    \begin{tabular}{c|cccccccccc}
    \toprule
    VLMs & AID & VRS-Cap & VRS-VG & VRS-VQA & SAR-VG & SAR-VQA & IR-VG & IR-VQA & FireRisk & RescueNet  \\ \midrule
    Qwen2.5-VL~\cite{Qwen2.5-VL}       & 63.3 & 0.3 & 40.8 & 62.5 & 36.6 & 44.9 & 53.4 & 47.7 & 14.5 & 43.3 \\
    MiniGPT-v2~\cite{chen2023minigpt}  & 24.5 & 2.4 & 0.3  & 40.5 & 0.8  & 32.8 & 0.1  & 27.0 & 14.8 & 29.3 \\
    LLaVA-1.5~\cite{liu2023visual}     & 48.0 & 1.5 & 1.9  & 57.3 & 1.0  & 39.8 & 0.2  & 38.1 & 16.1 & 38.1 \\
    GeoChat~\cite{kuckreja2024geochat} & 73.5 & 3.2 & 6.2  & 54.1 & 4.4  & 39.7 & 0.1  & 33.9 & 14.3 & 37.3 \\
    VHM w/o SFT~\cite{pang2025vhm}     & 34.6 & 0.0 & 0.2  & 46.9 & 0.2  & 49.5 & 0.0  & 25.5 & 14.3 & 23.6 \\ \bottomrule
    \end{tabular}
   }
\end{table}

\begin{table}[ht!]
    \centering
    \caption{Performance of VHM on CLeaRS (modality-incremental) with frozen vs. unfrozen vision encoder. $^\dagger$ indicates an unfrozen vision encoder. \textit{Inst.} reports performance immediately after each subset, while \textit{Final} denotes performance after the last subset, both under sequential fine-tuning.}
    \label{tab:vision-encoder-modality}
    \resizebox{\textwidth}{!}{
    \begin{tabular}{cc|cccccc|cccc}
    \toprule
    \multicolumn{2}{c|}{VLMs} & VRS-VG & SAR-VG & IR-VG & VRS-VQA & SAR-VQA & IR-VQA & MFT & MFN & MAA & BWT \\ \midrule
    \multirow{2}{*}{VHM} & Inst. & 5.7 & 12.7 & 12.6 & 72.1 & 91.9 & 56.0 & \multirow{2}{*}{41.8} & \multirow{2}{*}{24.2} & \multirow{2}{*}{15.7} & \multirow{2}{*}{-21.2} \\
     & Final & 0.7 & 1.2 & 0.3 & 44.3 & 42.5 & 56.0 & & & & \\ \midrule
    \multirow{2}{*}{VHM$^\dagger$} & Inst. & 8.7 & 33.6 & 37.3 & 72.7 & 92.4 & 58.8 & \multirow{2}{*}{50.6} & \multirow{2}{*}{26.3} & \multirow{2}{*}{19.1} & \multirow{2}{*}{-29.1} \\
     & Final & 0.1 & 0.0 & 0.0 & 32.1 & 66.8 & 58.8 & & & & \\
     \bottomrule
    \end{tabular}
    }
\end{table}

\subsection{Why is Visual Grounding Challenging in CLeaRS?}
As shown in Table 2 of the main paper, several models struggle to learn visual grounding tasks. During sequential fine-tuning, their performance on VRS-VG, SAR-VG, and IR-VG remains low even when evaluated immediately after training on the corresponding subset. Following prior works~\cite{zhang2024earthgpt,zhang2024popeye} on adapting vision-language models for multimodal remote sensing interpretation, we freeze the vision encoder throughout the sequential fine-tuning process. This choice may limit the model's ability to learn visual grounding tasks. 

To examine this factor, we unfreeze the vision encoder of VHM and perform sequential fine-tuning under the modality-incremental setting. As shown in Table~\ref{tab:vision-encoder-modality}, unfreezing the vision encoder leads to clear performance improvements on VRS-VG, SAR-VG, and IR-VG, resulting in higher MFT and MFN compared with the frozen-encoder variant. However, allowing the vision encoder to update also causes the model to focus more on the current subset during sequential fine-tuning, which increases forgetting of previously learned visual concepts. Consequently, more severe forgetting is observed, with BWT dropping from -21.2 to -29.1.

\end{CJK*}
\end{document}